\documentclass[lettersize,journal]{IEEEtran}

\usepackage{hyperref}
\usepackage{graphicx}
\usepackage{amsmath,amssymb,amsfonts}
\usepackage{cite}
\usepackage{gensymb}
\usepackage[caption=false,font=footnotesize]{subfig}

\usepackage{xcolor}

\usepackage{textcomp}
\usepackage{stfloats}
\def\BibTeX{{\rm B\kern-.05em{\sc i\kern-.025em b}\kern-.08em
    T\kern-.1667em\lower.7ex\hbox{E}\kern-.125emX}}
\usepackage{balance}

\usepackage{float}
\usepackage{tabularx}
\usepackage{booktabs}
\usepackage{arydshln}
\usepackage{longtable}
\floatstyle{plaintop}
\restylefloat{table}

\usepackage{titlesec}
\usepackage{dashrule}
\usepackage{pgf-pie}
\usepackage{multirow}

\newsavebox\tmpbox
\newcolumntype{G}[1]{>{\raggedright\let\newline\\\arraybackslash\hspace{0pt}}m{#1}}
\newcolumntype{C}[1]{>{\centering\let\newline\\\arraybackslash\hspace{0pt}}m{#1}}
\newcolumntype{D}[1]{>{\raggedleft\let\newline\\\arraybackslash\hspace{0pt}}m{#1}}

\begin{document}


\title{Experimental investigation of a maneuver selection algorithm for vehicles in low adhesion conditions}


\author{ Olivier Lecompte, William Therrien and Alexandre Girard
\thanks{Olivier Lecompte, William Therrien and Alexandre Girard are with the Department of Mechanical Engineering, Universite de Sherbrooke, Qc, Canada }
\thanks{\textcopyright IEEE. Personal use of this material is permitted. Permission from IEEE must be obtained for all other uses, in any current or future media, including reprinting/republishing this material for advertising or promotional purposes, creating new collective works, for resale or redistribution to servers or lists, or reuse of any copyrighted component of this work in other works. DOI:10.1109/TIV.2022.3188942}
}%

\markboth{IEEE Transactions on Intelligent Vehicles,~Vol.~7, No.~4, September~2022, Preprint version. DOI:10.1109/TIV.2022.3188942
}%
{}

\maketitle

\begin{abstract}
Winter conditions, characterized by the presence of ice and snow on the ground, are more likely to lead to road accidents. This paper presents an experimental proof of concept, with a 1/5th scale car platform, of a maneuver selection scheme for low adhesion conditions. In the proposed approach, a model-based estimator first processes the high-dimensional sensors data of the IMU, LIDAR and encoders to estimate physically relevant vehicle and ground conditions parameters such as the inertial velocity of the vehicle $v$, the friction coefficient $\mu$, the cohesion $c$ and the internal shear angle $\phi$. Then, a data-driven predictor is trained to predict the optimal maneuver to perform in the situation characterized by the estimated parameters. Experimental results show that it is possible to 1) produce a real-time estimate of the relevant ground parameters, and 2) determine an optimal maneuver based on the estimated parameters between a limited set of maneuvers.
\end{abstract}

\section{Introduction}


\IEEEPARstart{T}{he} rise of intelligence in vehicle mobility led to the implementation of several advanced driver-assistance systems (ADAS) in modern vehicles. Among these are the anti-lock braking system, the autonomous emergency braking and even combined control of steering and braking in situations where braking alone is not enough to avoid a collision \cite{safety_distance, hayashi_autonomous_2012}. However, when evolving in winter conditions, their performance is significantly reduced \cite{wang_development_2021}. According to the 2019 Road Safety Record of the Society of automobile insurance of Quebec, since the last 6 years, the number of total accidents goes up by around 30 \% during months featuring winter road conditions \cite{bilan_2019_saaq}. Encountering snow and ice leads to perception challenges \cite{yadav_real-time_2021} but also makes it more complex to predict the behavior of a vehicle and plan accordingly. 
%
%
An opportunity for safer vehicles would be to develop ADAS or autonomous driving systems that adapt their behavior according to the road conditions \cite{subosits_racetrack_2019, premachandra_smooth_2019}. This paper explores the opportunity of optimizing maneuver selections according to estimated ground parameters.

\begin{figure}[H]
	\centering
		\includegraphics[width=1\linewidth]{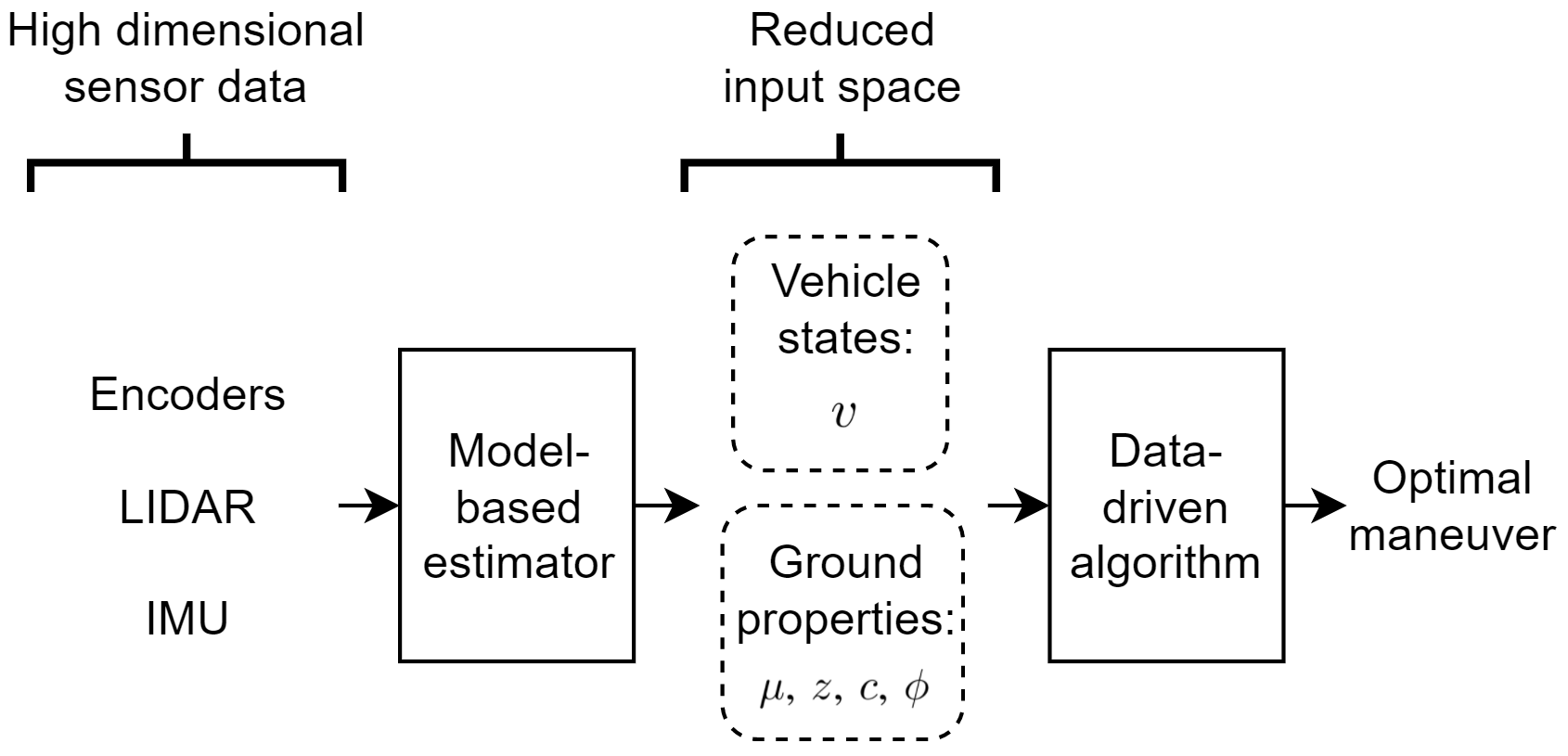}
		\vspace{-20pt}
	\caption{Proposed approach for the maneuver selection algorithm.}
	\label{fig:schema_projet}
\end{figure}

With a model-based approach, for instance model predictive control \cite{cheng_longitudinal_2020, brown_coordinating_2020}, one option would be to continuously estimate ground parameters and update them in the internal model of the controller. However, tire dynamics on ice, water and snow covered roads are difficult to model \cite{lee_vehicle-wet_2016, southwell_terramechanics_nodate}  and it would be very hard to integrate in a real-time controller. At the other end of the spectrum, with a data-driven approach, since the sensor signals are high-dimensional, an end-to-end approaches would require an unrealistically huge amount of data in this context. The evaluated approach in this paper is a hybrid  addressing these limitations. As illustrated in Fig. \ref{fig:schema_projet}, it is proposed here to use a model-based estimator providing a reduced input-space, characterizing the vehicle-road interaction, to a data-driven algorithm trained to select the optimal maneuver to perform.

\begin{figure}[H]
	\centering
        \includegraphics[width=0.55\linewidth]{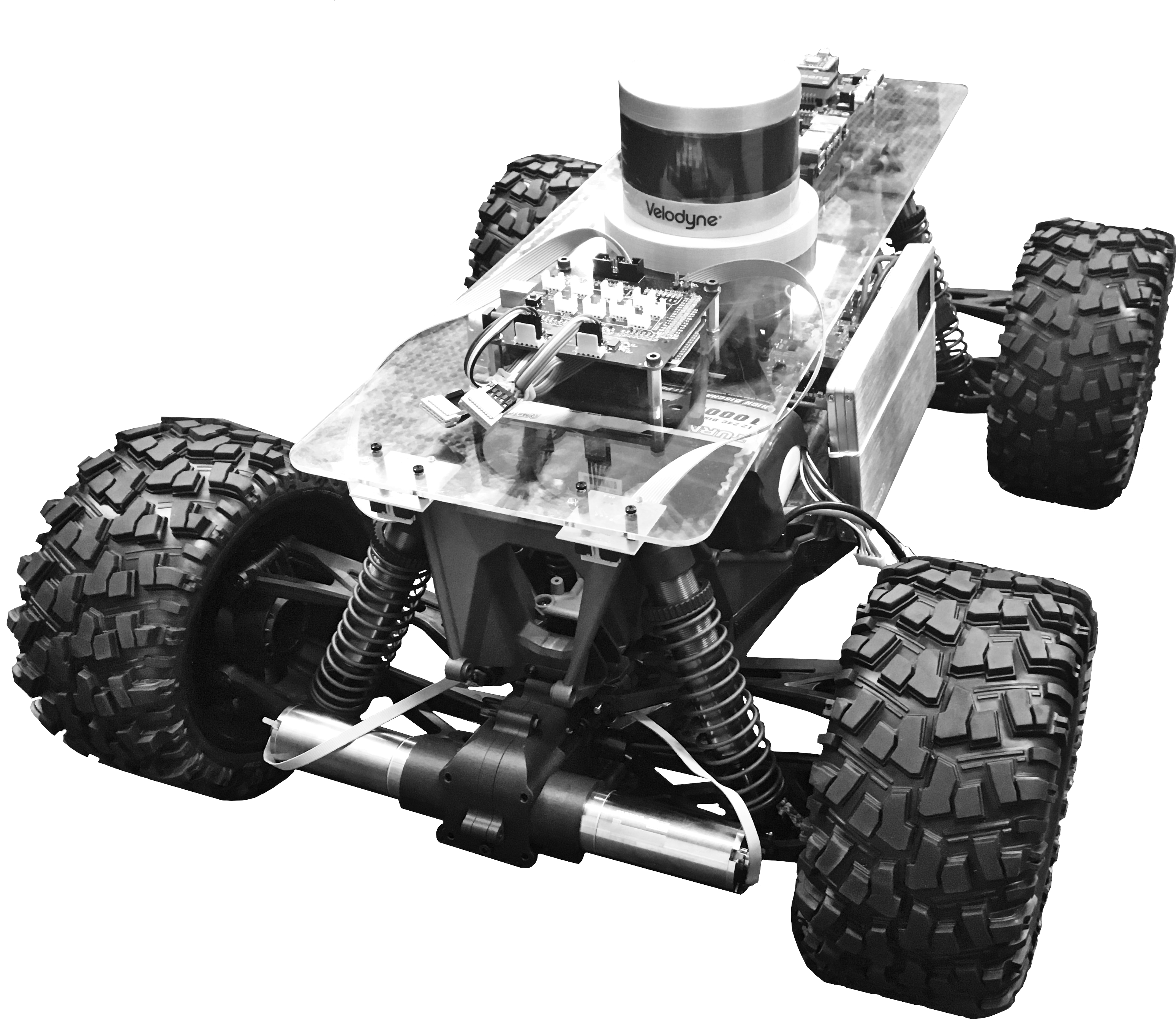} 
		\vspace{-7pt}
	\caption{Photo of the modified Traxxas X-MAXX used for the experimentation.}
	\label{fig:vehicle}
\end{figure}

This paper presents an evaluation of the potential of the proposed approach with an experimental proof-of-concept. To do so, a simplified case study has been established with a 1:5 reduced scale platform, illustrated in Fig. \ref{fig:vehicle}, in order to validate the feasibility of 1) estimating in real-time relevant ground parameters and 2) using those parameters to select an advantageous maneuver with a data-driven selector. 

The chosen case study is a collision avoidance situation where the vehicle is initially at a constant velocity towards a static obstacle located at a distance of 6 m. The objective is defined as selecting a maneuver, in a set of five discrete options, in order to maximize the minimum distance between the obstacle and the resulting car trajectory. Section \ref{sec:background} discusses the relevant related works, section \ref{sec:estimator} presents the algorithms that compose the model-based estimator, section \ref{sec:predictor} details the data-driven algorithm and section \ref{sec:exp} presents and discusses the experimental results.


\section{Background}\label{sec:background}
\noindent Some maneuver selection strategies incorporate  a minimal value or an estimation of the coefficient of friction in the decision making process \cite{fors_autonomous_2021, sevil_development_2019}. Several estimation techniques are proposed in the literature. Among them are the slip-slope method, which uses the relationship between the forces applied on the vehicle and the wheel slip \cite{rajamani_tire-road_2010}, various methods based on the longitudinal and lateral dynamics of the vehicle \cite{ahn_robust_2012} and some works even discuss the use of complementary information, such as image processing, acoustic sensors and ambient temperature measurement \cite{koskinen_sensor_2010}. 

The more complex characterization of the interaction between a vehicle and a deformable terrain, for instance sand, gravel or snow, has been studied especially in the context of rover explorations \cite{iagnemma_online_2004}. A model-based estimation algorithm was proposed to optimize torque inputs\cite{iagnemma_online_2004}. Also, some work proposes classifying different types of terrain by retrieving terramechanics properties with data-driven approaches \cite{southwell_terramechanics_nodate}. However, state-of-the-art maneuver selection systems rarely leverage ground properties since they are not developed for deformable terrain conditions. 

Recently, several artificial intelligence algorithms have been proposed for maneuver selection strategies. Reinforcement learning was used to decide how a vehicle should brake or turn in order to minimize the risk of accidents with an obstacle \cite{fu_decision-making_2020, arvind_autonomous_2019}. Some algorithms merge camera vision with ultrasonic sensors to address the problems of perception that are generated by the presence of snow and fog related to winter conditions \cite{yadav_real-time_2021}.  Many algorithms have been proposed to address perception issues in situations where the behavior is purely kinematic, but there has been much less investigation of maneuver selection in low adhesion conditions \cite{wang_development_2021}.

The contribution of the presented work is a demonstration of a novel data-driven maneuver selection algorithm leveraging estimated information characterizing ground properties to optimize the selected maneuver in low adhesion conditions.

\label{sec:rel}

\section{Model-based estimator} \label{sec:estimator}

\subsection{Determination of the relevant parameters} \label{subsec:param}

\begin{figure}[H]
\begin{center}
\includegraphics[width=0.9\linewidth]{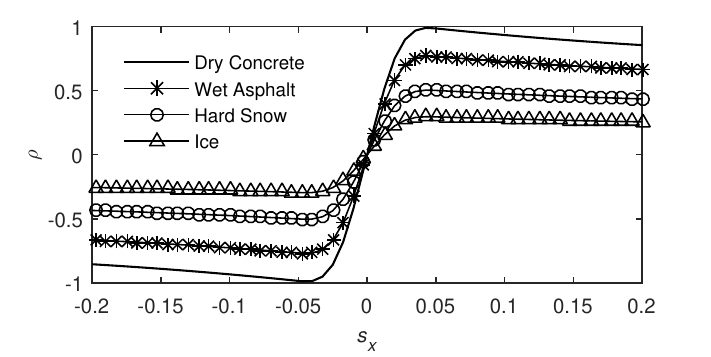}
\caption{Illustration of the slip-slope relation for typical flat grounds.}\label{fig:slipslope}
\end{center}
\end{figure}

On a hard surface, the relationship between the vehicle and the road is characterized by the friction coefficient $\mu$ \cite{rosenberger_dynamics_2016}. The latter is a function of the normalized traction force $\rho$ and the slip-ratio $s_x$. Both terms are defined in equation \eqref{eq:slip}, where $m$ is the mass of the vehicle, $g$ is the gravitational constant, $v$ is the inertial velocity of the vehicle, $\omega$ is the angular velocity of the wheel and $R$ is the radius of the wheel. The slip-slope, illustrated at Fig. \ref{fig:slipslope}, is defined as the relation between $s_x$ and $\rho$. For values of $s_x$ between $\pm 0.03$, the literature informs that $\mu$ is proportional to the slope of the relation whereas for large values of $s_x$, $\mu$ corresponds to the plateau of the $\rho(s_x)$ curve \cite{rajamani_tire-road_2010}.

\begin{equation} \label{eq:slip}
    \rho = \frac{F_x}{F_z} = \frac{m a}{m g}; \hspace{0.5cm} s_x = \frac{(\omega R - v)}{\max(\omega R, v) }
\end{equation}

However, in the presence of deformable terrain, an accumulation of particles is likely to form in front of the wheel. The interface between the wheel and the terrain generates a stress region which affects the behavior of the vehicle. The wheel sinkage $z$, the cohesion $c$ and the internal shear angle $\phi$ of the soil are parameters typically used to characterize the wheel-ground behavior in such situations \cite{iagnemma_online_2004}. 

In this proof of concept, the simplified emergency situation is defined as a vehicle encountering an obstacle of negligible width (like a post) that is detected at a distance of 6 m, on a ground that is flat and level. These features are chosen to emulate an emergency situation. The varying conditions that are studied are the initial vehicle velocity and the type of ground. Those will be described here by the set of parameters presented in table \ref{tab:variables}. Two parameters, $[ v, \mu ]$, are used in the case of the hard surface and four parameters, $[v, z, c,\phi ]$, are used for the ground with accumulation. 

\begin{table}[thpb]
  \caption{Presentation of the variables.} 
  \label{tab:variables}
  \centering 
  \begin{tabular}{C{1.5cm}C{1.75cm}G{4cm}} 
  \toprule[\heavyrulewidth]\toprule[\heavyrulewidth]
  \textbf{Symbol} & \textbf{Unit} & \textbf{Description} \\ \midrule
  \multicolumn{3}{c}{{Relevant parameters for the emergency maneuver selection}}\\ \midrule
  $v$ & m/s & Vehicle speed \\
  $\mu$ & --- & Friction coefficient \\
  $z$ & m & Wheel sinkage \\
  $c$ & kPa & Cohesion \\
  $\phi$ & rad & Internal shear angle \\ \midrule
  \multicolumn{3}{c}{{Intermediate variables}}\\ \midrule
  $T$ & Nm & Motor torque \\ 
  $s_x$ & --- & Longitudinal slip \\
  $\rho$ & --- & Normalized traction force \\
  $\omega$ & rad/s & Wheel angular velocity \\
  $a$ & m/s$^2$ & Vehicle linear acceleration \\ 
  \bottomrule[\heavyrulewidth] 
  \end{tabular}
\end{table}

\subsection{Estimation algorithms}
\begin{figure}[thpb]
\begin{center}
\includegraphics[width=1\linewidth]{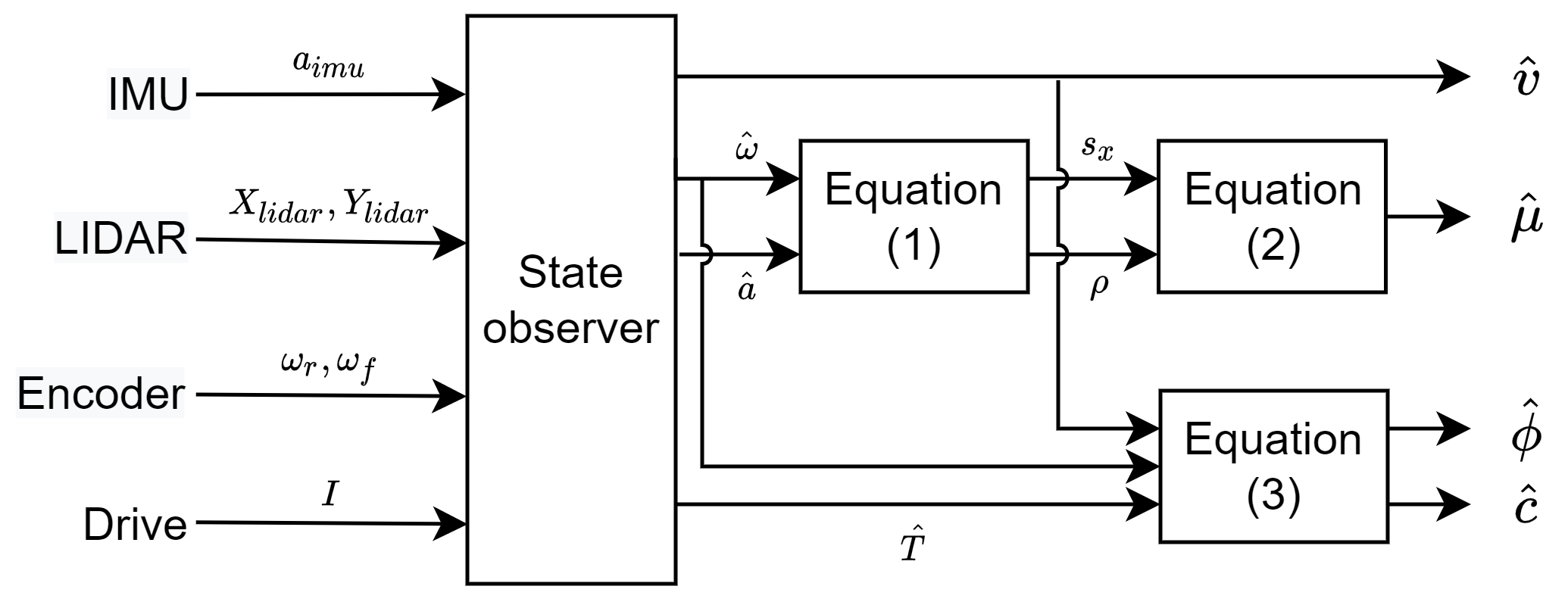}
\caption{Illustration of the model-based estimator.}\label{fig:estimator_schema}
\end{center}
\vspace{-10pt}
\end{figure}

The model-based estimator, illustrated at Fig. \ref{fig:estimator_schema}, is composed of three estimation algorithms, one for the vehicle states $v$, $\omega$ and $a$, one for the friction coefficient $\mu$ and one for the terramechanics properties $c$ and $\phi$. 

\subsubsection{Vehicle states estimation}
A state observer outputs an estimate of the velocity $\hat{v}$, the vehicle acceleration $\hat{a}$ and the front wheels angular velocity $\hat{\omega}$ based on the sensor signals $[ X_{lidar}, Y_{lidar} ]$ provided by the LIDAR (with a SLAM algorithm), $\omega_r$ measured by the rear wheels encoders, $\omega_f$ provided by the front wheel encoders, $a_{imu}$ provided by the IMU and an estimate of the motor torque $\hat{T}$ obtained from a measurement of the current $I$ provided by the drive. The observer used is based on a linear dynamic model identified with experimental data and its gains were adjusted to obtain an adequate compromise between tracking performance and output noise. 

\subsubsection{Friction coefficient} \label{metho:friction}
The estimation algorithm for the friction coefficient is based on the slip-slope relation presented at section \ref{subsec:param}. Here, only the maximum traction force $\rho$ is estimated, corresponding to the plateau of the curve at Fig. \ref{fig:slipslope}. First, $s_x$ and $\rho$ are computed according to equation \eqref{eq:slip}, using $\hat{v}$, $\hat{\omega}$ and $\hat{a}$. Then, the estimation $\hat{\mu}$ is updated based on a sliding average of computed $\rho$ in the last N time steps while $s_x \geq 0.03$:
\begin{equation} \label{eq:moy_gliss}
\hat{\mu} = \begin{cases}
\hat{\mu}_{old}+ \frac{1}{N} ( |\rho_{n=1}| - |\rho_{n=N}|) & \text{if $|s_x| \geq 0.03$}\\
\hat{\mu}_{old} &\text{if $|s_x| < 0.03$}
\end{cases}
\end{equation}
where $\hat{\mu}_{old}$ is the last estimated value, $\rho_{n=1}$ is the newest normalized traction force calculation and $\rho_{n=N}$ the oldest value of the sliding window. 

\subsubsection{Cohesion, internal friction angle and sinkage} \label{metho:terra}
In the presence of deformable terrain, instead of computing the friction coefficient, an other
estimation algorithm is used to obtain an estimate of the cohesion $c$ and the internal shear angle $\phi$ that characterize the ground. For this purpose, the linear least-squares estimation method proposed by Iagnemma et al. \cite{iagnemma_online_2004} is used as follow: 
\begin{equation} \label{eq:iagnemma}
    \begin{bmatrix}
    \hat{c} \\
    \tan \hat{\phi}
    \end{bmatrix} = (K_2^T K_2)^{-1} K_2^T K_1
\end{equation}
where matrices $K_1$ and $K_2$ are composed of $j$ data points of estimated coefficients:
\begin{equation*}
    K_1 = \begin{bmatrix}
    \frac{\kappa_2^1}{\kappa_3^1} & ... & \frac{\kappa_2^j}{\kappa_3^j}
    \end{bmatrix}^T \hspace{0.5cm}; \hspace{0.5cm}
    K_2 = \begin{bmatrix}
    1 & ... & 1 \\
    \frac{-\kappa_1^1}{\kappa_3^1} & ... & \frac{-\kappa_1^j}{\kappa_3^j}
    \end{bmatrix}^T
\end{equation*}
defined as follows:
\begin{align*}
    \kappa_1 & = \alpha\bigg(\beta^2m g R+4 T\sin\beta_1-8 T\sin\frac{\beta}{2} \bigg) \\
    \kappa_2 & = 4T \bigg(\cos\beta-2\cos\frac{\beta}{2}+1 \bigg) \\
    \kappa_3 & = \beta R^2b \bigg(\cos\beta-2\cos\frac{\beta}{2} \\ & +2\alpha\cos\beta-4\alpha\cos\frac{\beta}{2}+2\alpha+1 \bigg)  
\end{align*}   
where $b$ is the width of the wheel and
\begin{align*}
    \alpha  & = 1-e^{\frac{R}{b} [ \frac{\beta}{2}+(1-\zeta)(-\sin\beta+\sin\frac{\beta}{2} ]}\\
    \beta &= \arccos{\frac{R-z}{R}} \\
    \zeta &= 1-1/\bigg( \frac{\hat{v}}{\hat{\omega R}}\bigg)
\end{align*}

In the presented work, the presence of a deformable layer is assumed to be known in advance and the sinkage $z$ is provided to the estimator. Future work will investigate adding instrumentation and an estimation algorithm to automatically characterize the sinkage in real-time.

\section{Data-driven predictor} \label{sec:predictor}
The data-driven predictor has the task of selecting the optimal maneuver to execute based on the reduced input-space produced by the model-based estimator. Since the dynamic behavior of a vehicle in low adhesion condition is hard to model analytically, this step is purely data-driven. 

\begin{figure}[thpb]%
    \centering
    \hspace{-12pt}
    \subfloat[]{{\includegraphics[height=0.3\linewidth]{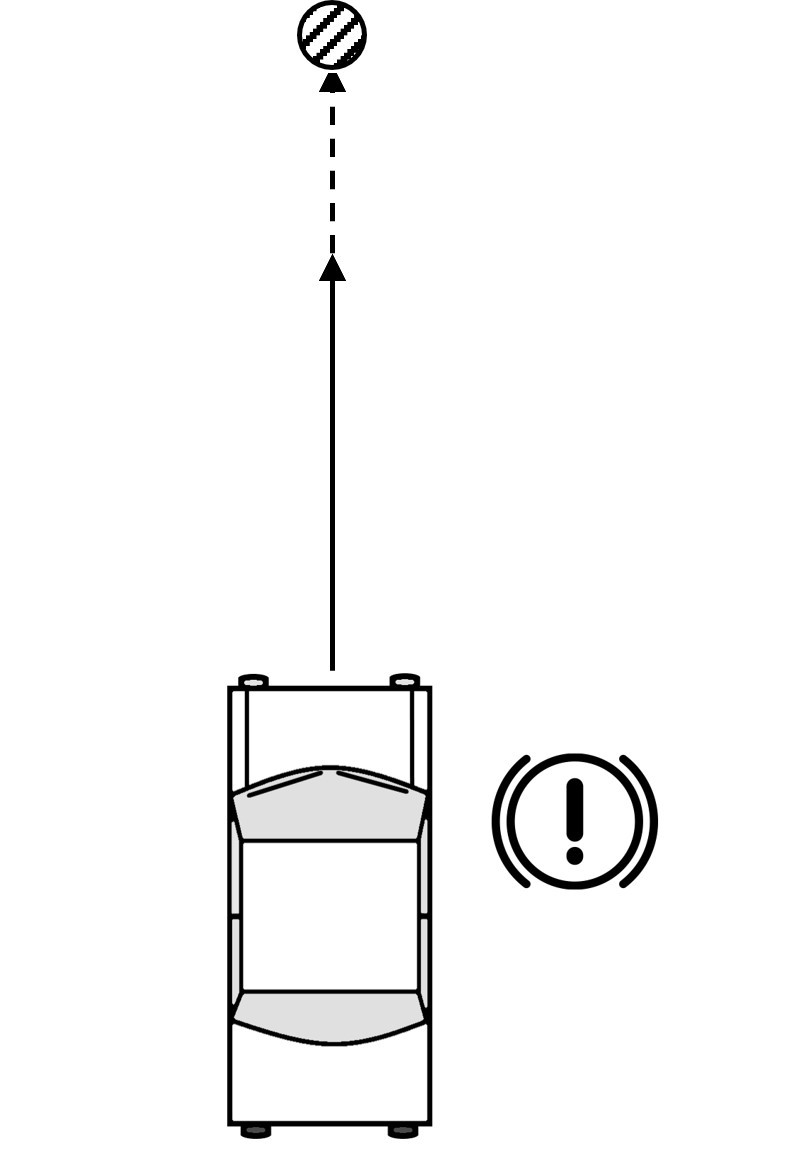} }} \label{fig:man1}
    \hspace{-12pt}
    \subfloat[]{{\includegraphics[height=0.3\linewidth]{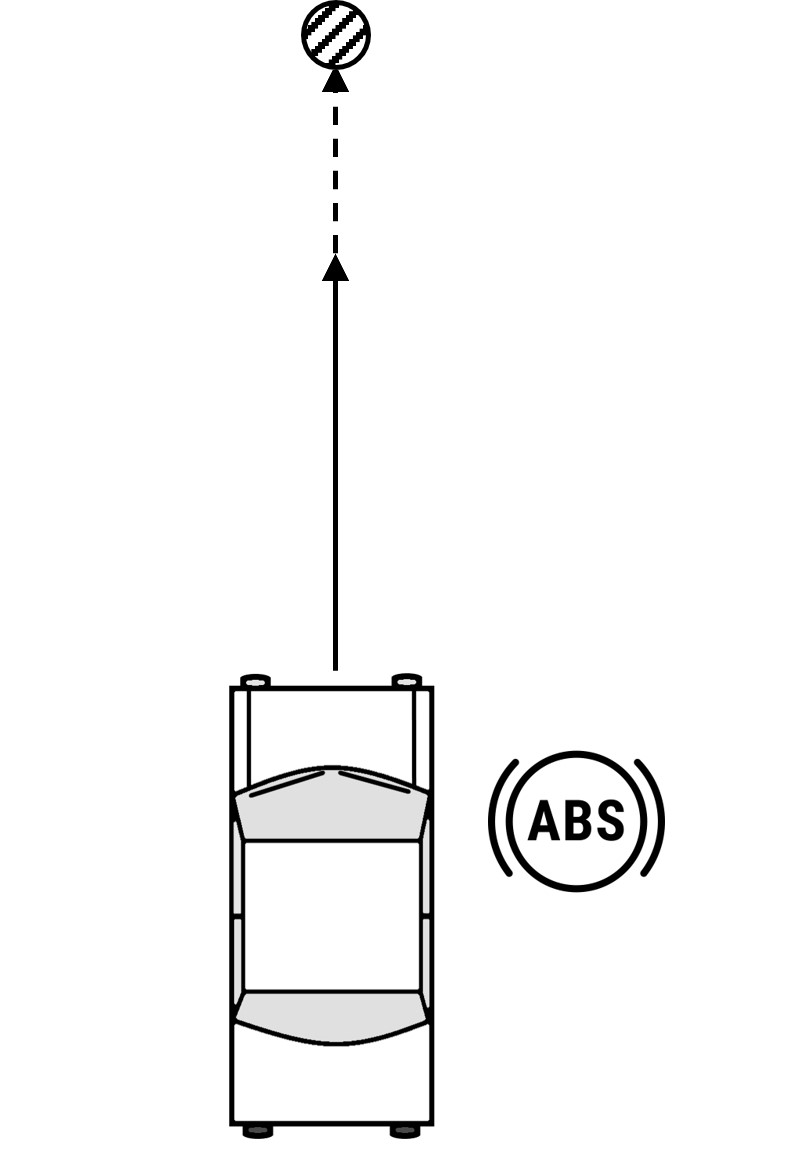} }} \label{fig:man.2}
    \hspace{-11pt}
    \subfloat[]{{\includegraphics[height=0.3\linewidth]{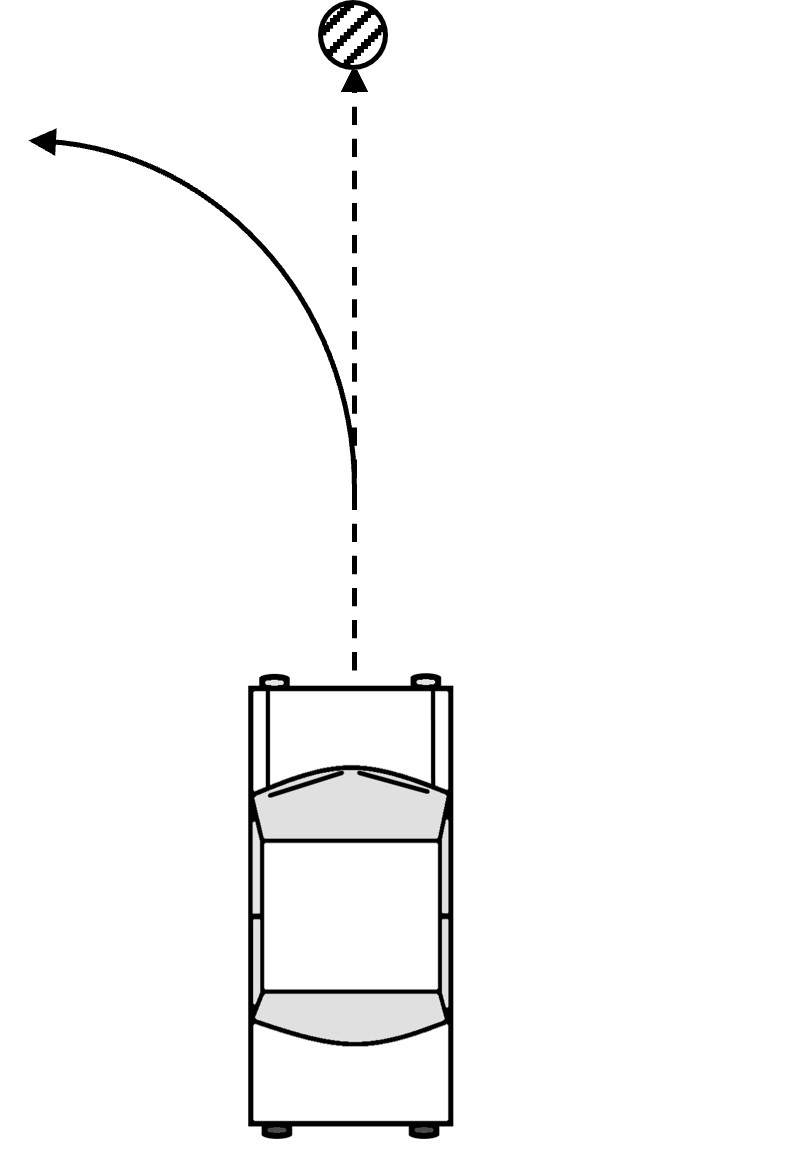} }} \label{fig:man.3}
    \hspace{-11pt}
    \subfloat[]{{\includegraphics[height=0.3\linewidth]{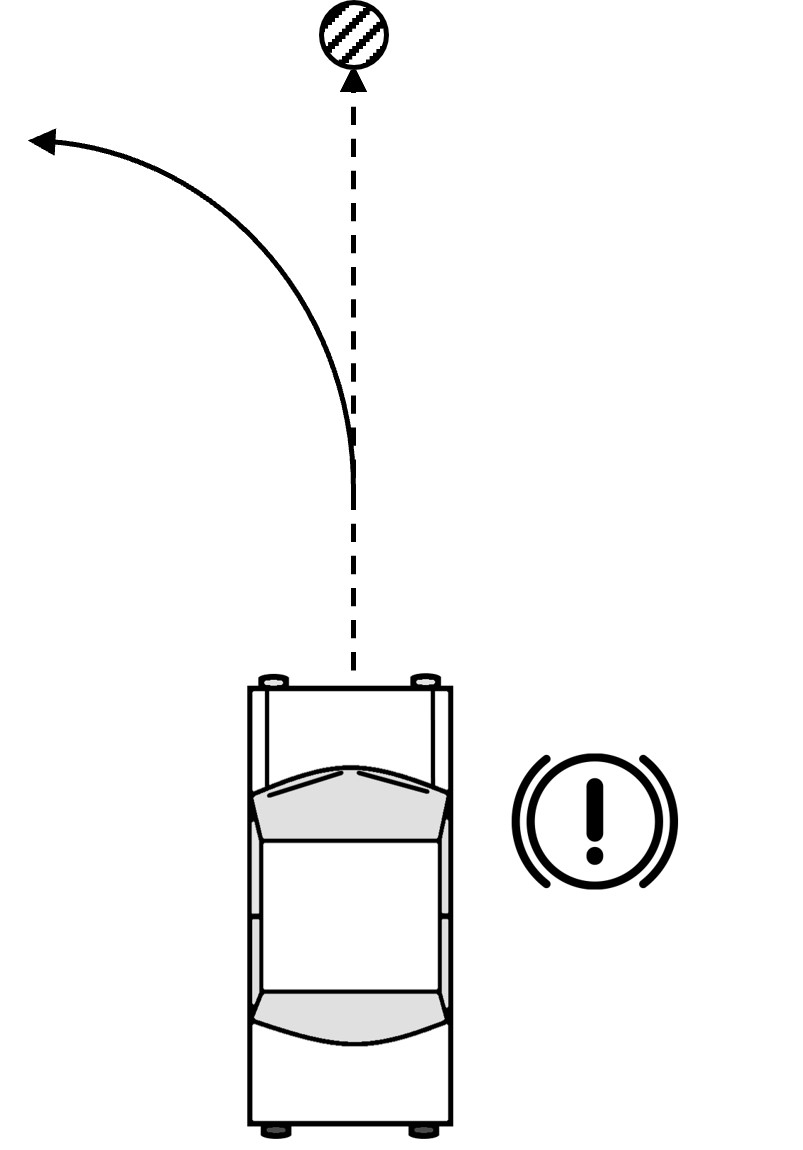} }} \label{fig:man.4}
    \hspace{-11pt}
    \subfloat[]{{\includegraphics[height=0.3\linewidth]{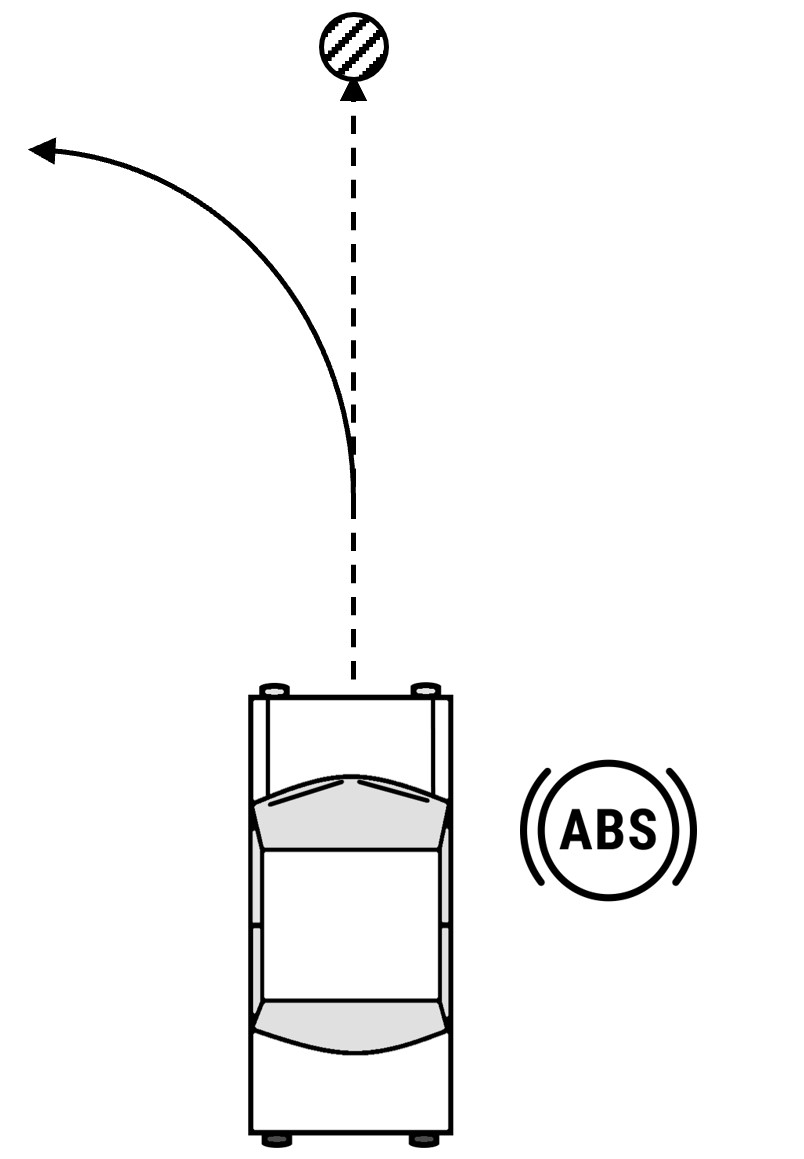} }} \label{fig:man.5}
    \caption{Illustration of the discrete set of maneuvers, where (1) is brake 100 \%, (2) is brake ABS, (3) is turn 100 \%, (4) is turn 100 \%, brake 100 \% and (5) is turn 100 \%, brake ABS.}%
    \label{fig:manoeuvers}%
\end{figure}

For this proof of concept a simplified emergency manoever context, that still capture a wide range of dynamic behavior, is used. The algorithm is developped to select a maneuver that maximizes the minimal distance $d$ between the resulting trajectory of the vehicle and an immobile punctual obstacle located at a distance of 6 meters in front of the vehicle. Five discrete maneuver options, illustrated in Fig. \ref{fig:manoeuvers}, are considered and are defined as follows:

\begin{enumerate}
    \item Brake 100 \% - The wheels are blocked;
    \item Brake ABS - The vehicle is braking with an anti-lock braking system (ABS);
    \item Turn 100 \% - The front wheels are turned to the maximal steering angle of 25 $\degree$. The propulsion commands is disabled;
    \item Turn 100 \%, Brake 100 \% - The front wheels are turned to the maximal steering angle while the wheels are blocked;
    \item Turn 100 \%, Brake ABS - The front wheels are turned to the maximal steering angle of 25 $\degree$ while the vehicle is braking with the ABS system.
\end{enumerate}

Note that here, the braking is implemented with a closed-loop velocity controller targeting $\omega_f = 0$ rad/s and the ABS is implemented by setting front motor currents to zero when $s_x>0.03$. In both instances, the rear wheels are free and braking is done only with the front wheels.

To select the optimal maneuver, first independent predictor functions compute the minimum distance $d$ for each maneuver option $i$. Linear regressions based on augmented input spaces are chosen for their simplicity and effectiveness. Two discrete functions are used, in the case of a hard surface ($z=0$) or in the case of deformable ground ($z>0$). The augmented input terms were selected by testing multiple options and keeping the one with the most influence on the predicted output. The predictor equations used are:   
\begin{equation} \label{eq:regression}
\hat{d}_i = \begin{cases}
\gamma_{0i} + \gamma_{1i} v + \gamma_{2i} \mu + \gamma_{3i} v \mu + \gamma_{4i} v^2 + \gamma_{5i} \mu^2 & \text{if $z=0$}\\
\lambda_{0i} + \lambda_{1i} v + \lambda_{2i} z + \lambda_{3i} v c + \lambda_{4i} \phi &\text{if $z>0$}
\end{cases}
\end{equation}
where $\gamma$ and $\lambda$ are the regression coefficients. Here, the regression is conducted using least squares on the 330 data points collected from the tests presented at section \ref{subsec:exp_seq}.
Finally, the optimal maneuver $i^*$, that is predicted to maximize this distance, is selected and executed by the vehicle:
\begin{equation}
i^*= \underset{i \in \{1,2,3,4,5 \}}{\operatorname{argmax}} \hat{d}_i
\label{eq:argmin}
\end{equation}

\section{Experimental validation}\label{sec:exp}

\subsection{Methodology}

\subsubsection{Material}
A  1:5 scale experimental vehicle platform based on a Traxxas X-MAXX is used to evaluate the proposed maneuver selection algorithm. All wheels are independent and controlled with 4 Maxon motors DCX32L. The relevant specifications are presented in table \ref{tab:specifications}. The on-board instrumentation is chosen to emulate the sensor sets present on modern intelligent vehicles. A SLAM algorithm is used to retrieve the location $[ X, Y]$ of the vehicle in space via the ROS package rtabmap.\cite{rtabmap_ros_2021}.

\subsubsection{Experimental sequence} \label{subsec:exp_seq}
To form the data-set required to train the data-driven predictors functions, 330 experimental tests are conducted to evaluate the result of each maneuver with multiple initial velocities $v$ \{1 m/s, 2 m/s, 3m/s\}, three types of hard ground (\ref{tab:surface_param}a), four types of ground with accumulation (\ref{tab:surface_param}b) and two values for $z$ \{1 cm, 3 cm\}.
%
%
%
For the ground with accumulation, custom mixes of sand, silt, fines and clay were used. Conforming to the ASTM-D2487 standard \cite{d18_committee_practice_nodate}, a poorly graded sand consists of 95~\% sand and 5~\% fines, a clayey sand consists of 88~\% sand and 12~\% clay and a clay loam consists of 52~\% sand, 40~\% silt and 8~\% clay. The theoretical cohesion and internal angle cohesion are taken from the literature \cite{user_cohesion_nodate, user_angle_nodate}. Every combination of vehicle initial velocity, type of ground and executed maneuver was conducted experimentally twice.

\begin{table}[h]
  \caption{Vehicle specifications} 
  \centering 
  \label{tab:specifications}
  \subfloat[Dimensions]{ \begin{tabular}{G{2.5cm}C{1.75cm}C{1.25cm}C{1.25cm}} 
  \toprule[\heavyrulewidth]\toprule[\heavyrulewidth]  
  \textbf{Dimension} & \textbf{Variable} & \textbf{Value} & \textbf{Unit}\\ \midrule
  Vehicle mass & m & 14.85 & kg \\
  Wheel radius & R & 0.1 & m \\
  Wheel width & b & 0.07 & m \\
  \bottomrule[\heavyrulewidth] 
  \end{tabular}}
  \qquad
  \subfloat[Sensors]{   \begin{tabular}{C{1.85cm}G{2.7cm}G{2.5cm}} 
  \toprule[\heavyrulewidth]\toprule[\heavyrulewidth] 
  \textbf{Measure} & \textbf{Description} & \textbf{Sensor}\\ \midrule
  $a_{imu}$ & Linear acceleration & IMU Xsens MTi-670  \\
  $X_{lidar}, Y_{lidar}$ & Vehicle position & 3D Lidar Velodyne PUCK\\
  $\omega_r, \omega_f$ & Wheel angular velocity & Encoder Maxon ENX10 \\
  \bottomrule[\heavyrulewidth] 
  \end{tabular}}
\end{table}

\begin{table}[thpb]
    \caption{Ground properties for the experimentation.} \label{tab:surface_param}
    \subfloat[Hard surface ground]{   \begin{tabular}{G{4.5cm}C{3.15cm}} \toprule[\heavyrulewidth]\toprule[\heavyrulewidth] 
    \textbf{Wheels and Ground composition} & \textbf{Friction coefficient $\mu$} \\ \midrule
   Plastic wheels on polyethylene & 0.25 \\
   Rubber wheels on linoleum & 0.45 \\
   Rubber wheels on rubber & 0.9 \\ \bottomrule[\heavyrulewidth] 
   \end{tabular}}
    \qquad
    \subfloat[Ground with accumulation]{   \begin{tabular}{G{3.75cm}C{1.5cm}C{2cm}} 
   \toprule[\heavyrulewidth]\toprule[\heavyrulewidth]  
   \textbf{Ground composition} & \textbf{Cohesion $c$ (kPa)} & \textbf{Internal friction angle $\phi$ (deg)}\\ \midrule
   Poorly graded sand & 0 & 35 \\
   Clayey sand (compacted) & 74 & 31 \\
   Clay loam (compacted) & 83 & 25 \\
   Clay loam (saturated) & 15 & 25 \\
   \bottomrule[\heavyrulewidth] 
   \end{tabular}}
\end{table}

\subsection{Results}

\subsubsection{Vehicle states estimation}
\begin{figure}[htpb]
\begin{center}
\includegraphics[width=.9\linewidth]{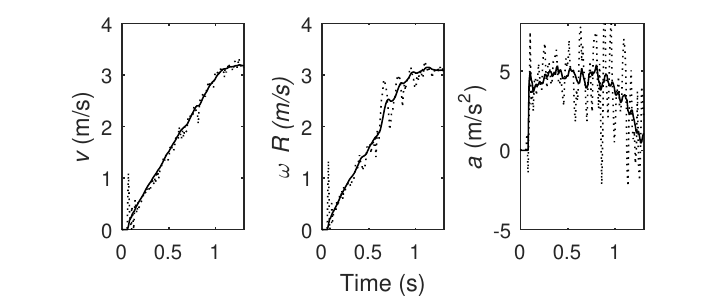}
\caption{State observer outputs for $v$, $\omega R$ and $a$ (---) and the corresponding raw signal data (\textperiodcentered\textperiodcentered\textperiodcentered).}\label{fig:observer_output}
\end{center}
\end{figure}

Fig. \ref{fig:observer_output} presents the estimated output values of the state observer, for an experimental sequence where the vehicle accelerates until a setpoint of 3 m/s is reached. The acceleration phase occurs on a hard ground characterized by a friction coefficient of $\mu = 0.45$ (rubber wheels on linoleum). 

\subsubsection{Friction coefficient}
\begin{figure}[thpb]
\begin{center}
\includegraphics[width=0.9\linewidth]{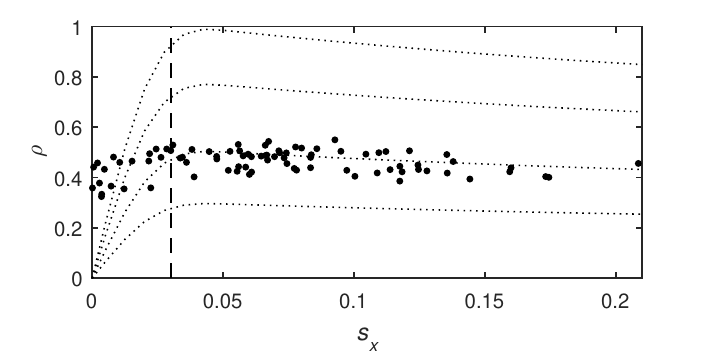}
\vspace{-10pt}
\caption{Superposition of the calculated ($s_x$, $\rho$) and typical curves.}\label{fig:slipslope_5_2}
\vspace{-10pt}
\end{center}
\end{figure}

\begin{figure}[thbp]
\begin{center}
\includegraphics[width=0.9\linewidth]{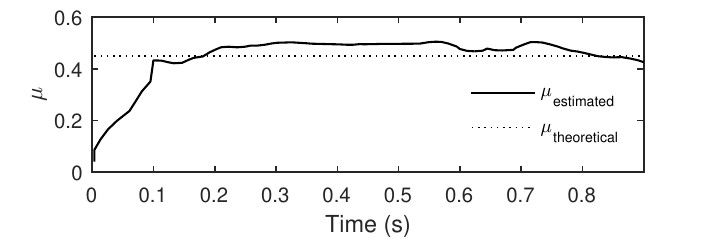}
\vspace{-10pt}
\caption{Estimation of $\mu$ in time, compared to the theoretical value.}\label{fig:mu_est_vs_t}
\vspace{-10pt}
\end{center}
\end{figure}

The same experimental sequence is used to evaluate the performance of the friction coefficient estimation. Fig. \ref{fig:slipslope_5_2} presents a superposition of the $( s_x, \rho)$ combinations calculated throughout the experimentation and the theoretical slip-slope relations for typical grounds. Points located to the right of the dashed line are used in the estimation of the friction coefficient as shown in equation \eqref{eq:moy_gliss}. Fig. \ref{fig:mu_est_vs_t} presents the output of the estimation algorithm for $\mu$ as a function of time for a sliding mean executed with $N = 10$ terms at 90 Hz. The estimated value converges to the theoretical value of 0.45 in about 0.1 seconds.

\subsubsection{Cohesion and internal friction angle}
\begin{figure}[H]
\begin{center}
\vspace{-10pt}
\includegraphics[width=.9\linewidth]{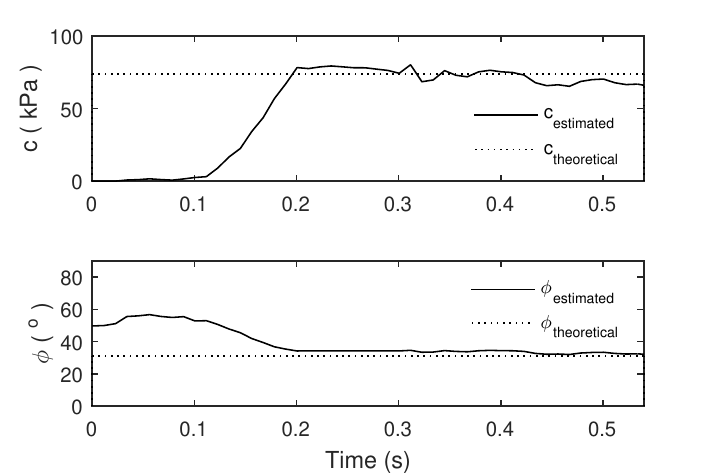}
\caption{Estimation of $c$ and $\phi$ in time, compared to the theoretical values.}\label{fig:c_phi_est}
\vspace{-10pt}
\end{center}
\end{figure}

An experimental sequence where the vehicle accelerates on a hard ground until a setpoint of 3 m/s is reached before decelerating by braking and blocking the wheels is conducted to evaluate the performance of the terramechanics properties estimator. The deceleration phase is performed on an accumulation of compacted clayey sand with $z = 0.03$ m. Fig. \ref{fig:c_phi_est} presents the output of the linear least-squares estimation algorithm for $c$ and $\phi$ with $j = 10$ data points. The estimated values converge rapidly near the theoretical values of 74 kPa for $c$ and 31 $\degree$ for $\phi$ in about 0.2 seconds.

\subsubsection{Data-driven predictor}
\begin{figure}[H] %
    \centering
    \vspace{-10pt}
    \subfloat[Optimal manoeuver on hard surface ground]{{\includegraphics[width=.75\linewidth,trim={0 1cm 0 0},clip]{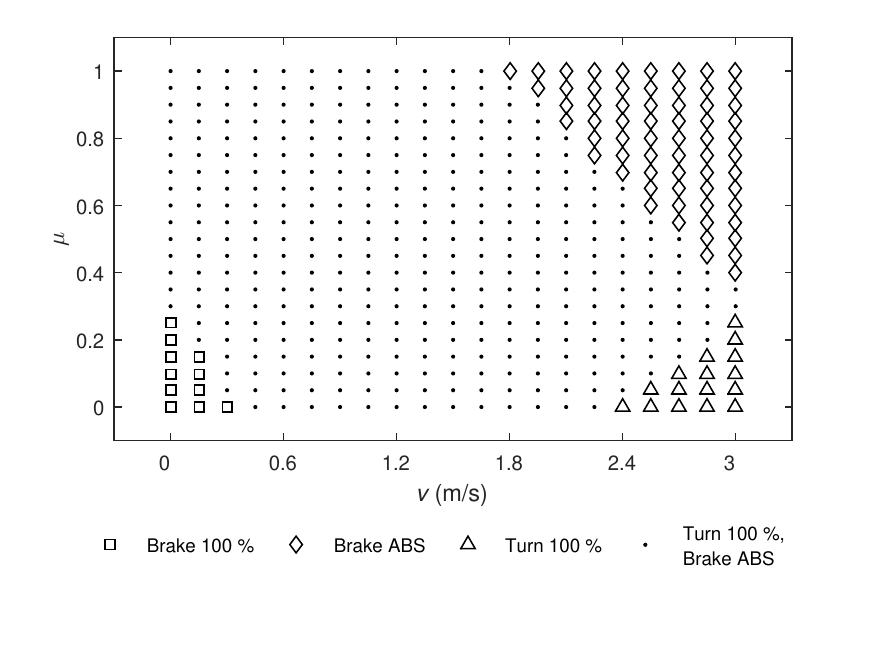} }}
    \vspace{-10pt}
    \qquad
    \subfloat[Optimal manoeuver on deformable terrain]{{\includegraphics[width=.75\linewidth,trim={0 1cm 0 0},clip]{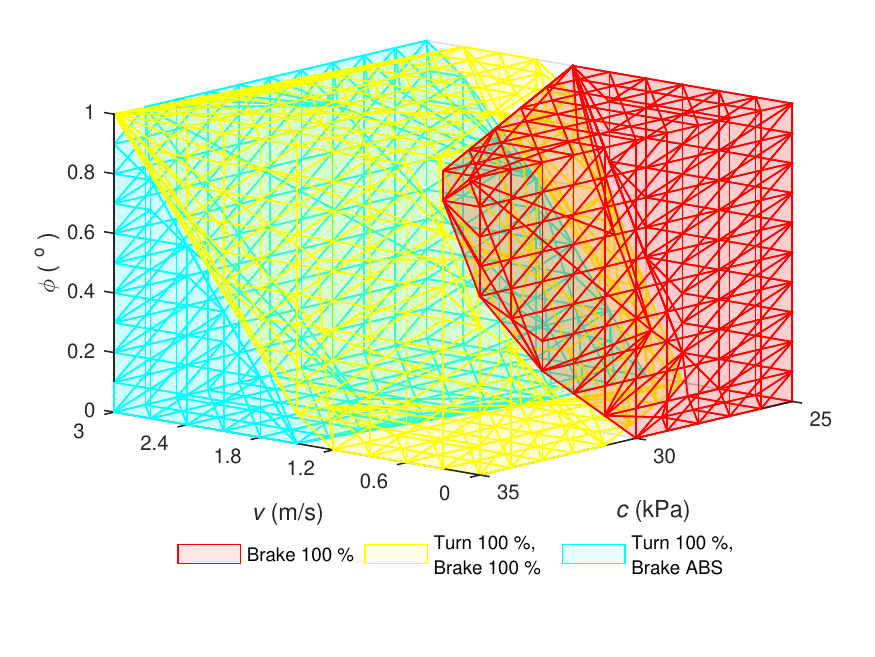} }}
    \caption{Decision maps for optimal maneuvers based on the physics-based parameters.}%
    \label{fig:predictor}
\end{figure}

Fig. \ref{fig:predictor} shows two decision maps indicating the optimal maneuvers to perform according to the trained algorithm (the output of equation \eqref{eq:argmin}). The optimal decision is displayed according to $[v, \mu]$ for the hard surface ground and according to $[v, c, \phi]$ for the ground with accumulation $z = 0.03$ m.

\begin{figure}[th]%
    \centering
    \subfloat[Maneuvers on a hard surface ground with $v = 3$ m/s and $\mu = 0.25$]{{\includegraphics[width=1\linewidth,trim={0 1cm 0 0},clip]{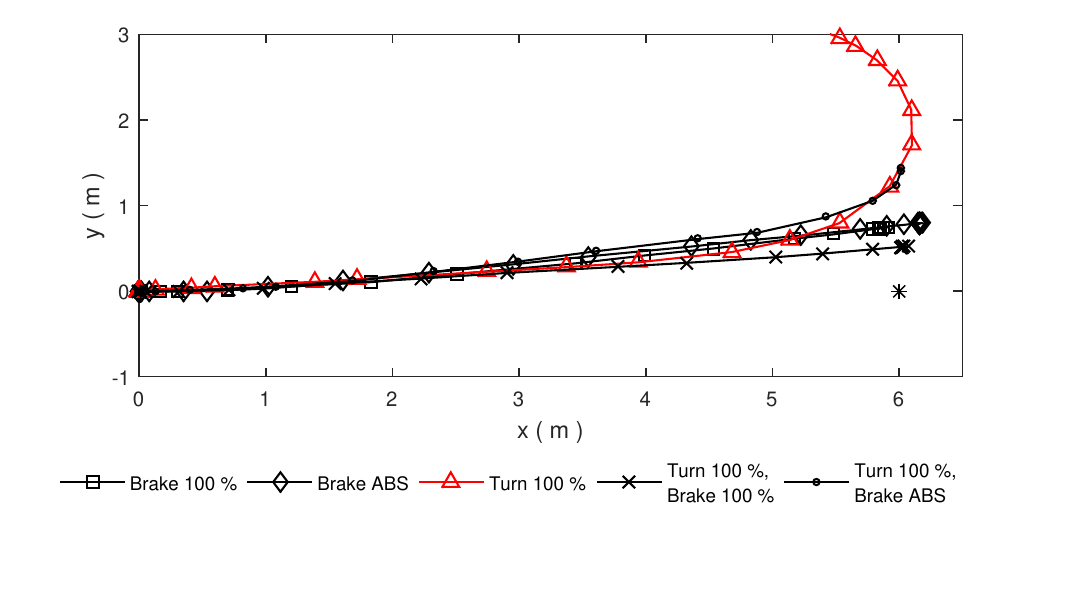} }} 
    \vspace{-10pt}
    \qquad
    \subfloat[Maneuvers on compacted clay loam with $v = 3$ m/s, and $z = 0.03$ m]{{\includegraphics[width=1\linewidth,trim={0 1cm 0 0},clip]{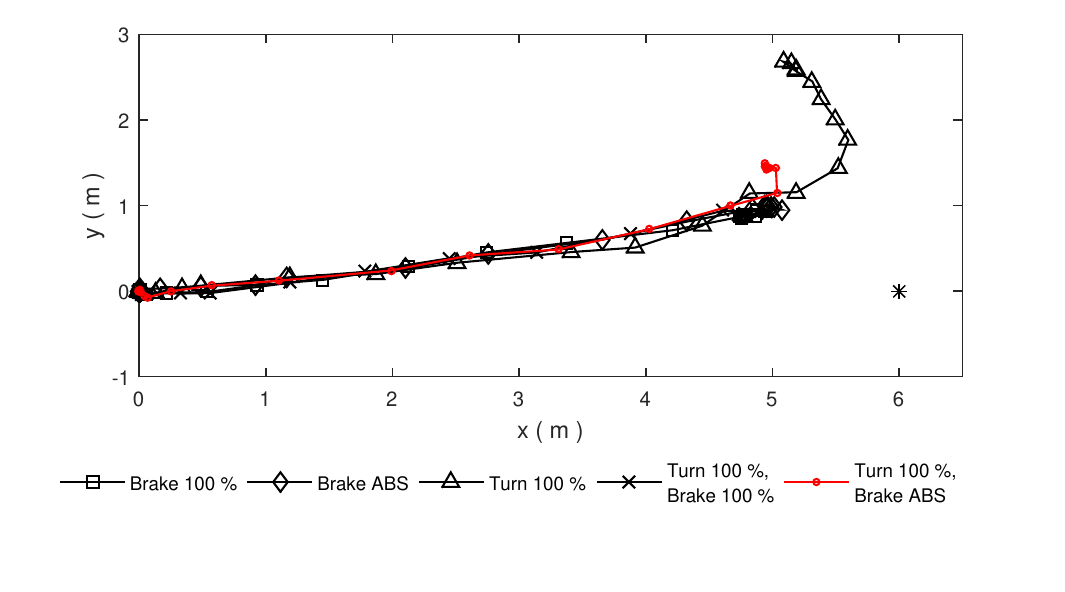} }}
    \caption{Illustration of the disparities in the optimal maneuvers for the different soil types. In both instances, the optimal maneuver is in red.}%
    \label{fig:manoeuvres_disparities}%
\end{figure}

\begin{figure}[thpb]%
    \centering
    \subfloat[Maneuvers on a hard surface ground with $v = 3$ m/s and $\mu = 0.25$]{{\includegraphics[width=1\linewidth]{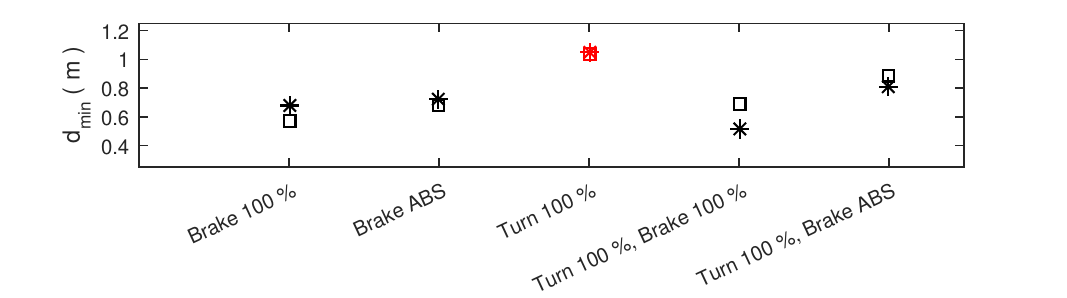} }}
    \vspace{-10pt}
    \qquad
    \subfloat[Maneuvers on compacted clay loam with $v = 3$ m/s, and $z = 0.03$ m]{{\includegraphics[width=1\linewidth]{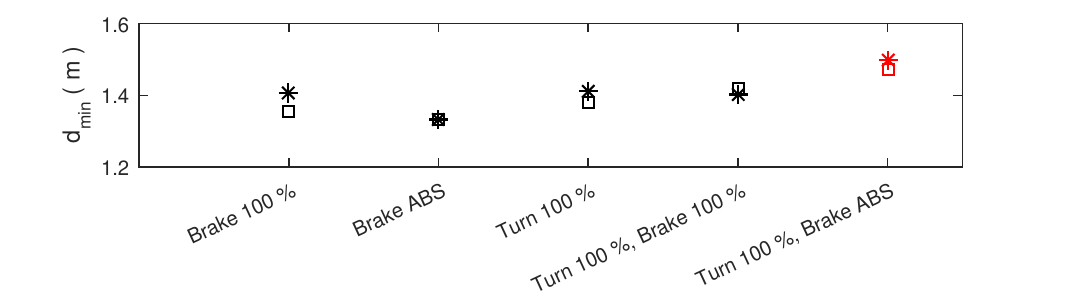} }}
    \caption{Comparison of the predicted values ($\Box$) and the experimental results ($*$). In both instances, the optimal maneuver is in red.}%
    \label{fig:precision_predictor}%
\end{figure}

Fig. \ref{fig:manoeuvres_disparities} presents the resulting trajectories of the 5 proposed maneuvers for a specific case: a vehicle initially at 3 m/s (1) for a hard surface ground for which $\mu = 0.25$ and (2) for a ground composed of compacted clay loam with $z = 0.03$ m. Also, for the same conditions, Fig. \ref{fig:precision_predictor} presents a preliminary visual assessment of the precision of the algorithm, showing the predicted minimum distance and the real minimum distance obtained during experimental sequences.

It can be seen that the presence of a deformable layer and its composition affects the behavior of the vehicle and that the predictor is able to select the optimal maneuver accordingly. Furthermore, for most conditions, the ABS braking is favorable on a hard surface but that locking the front wheel is preferable in the presence of a deformable layer. This seems to confirm the previous statements made in section \ref{subsec:param} in that the accumulation of particles in front of the locked wheel helps slowing the vehicle faster. This demonstrates the relevance of considering ground properties in the development of maneuver selection algorithms and shows the potential of the presented approach, despite a limited data-set.

\section{Conclusion and outreach}
\noindent To conclude, this paper presents a proof of concept for a hybrid model-based data-driven maneuver selection algorithm designed for low adhesion conditions. A model-based estimator provides the real-time estimation of a set of relevant parameters characterizing the interaction between a vehicle and the ground to allow a data-driven predictor to determine the optimal emergency maneuver to perform.

The results presented in this paper shows the relevance of using ground parameters to select the optimal maneuver in low adhesion situation, and the potential of the presented hybrid approach to improve collision avoidance algorithms. Since results were obtained in a simplified situation, in future work, it would be relevant to evaluate more complex emergency situations with a model-based estimator for a more comprehensive set of relevant parameters, for instance the obstacle position, its size, the ground inclination, the temperature, etc.. In addition, it would be interesting to test online learning where the data-driven predictor is continuously updated. Finally, it would be relevant to replace the discrete set of five possible maneuvers studied by a continuous action space (more freedom in the target trajectory) that could address the needs of more specific emergency situations.


\bibliographystyle{IEEEtran}
\bibliography{bib.bib}

\end{document}